%% file: main.tex
\documentclass[a4paper,10pt,conference]{ieeeconf}      

\IEEEoverridecommandlockouts                              
\usepackage{amssymb}
\usepackage{amsmath}
\usepackage{bm}
\usepackage{epsfig}
\usepackage{here}
\usepackage{flushend}
\usepackage{graphicx}
\usepackage{graphics}

\def\mR{\mathbb{R}}

\def\bAs{\mbox{\scriptsize{\boldmath$A$}}}
\def\rme{\mbox{${\rm e}$}}

\def\dth{\mbox{$\dot{\theta}$}}

\def\bp{\mbox{\boldmath$p$}}

\def\bW{\mbox{\boldmath$W$}}

\def\bA{\mbox{\boldmath$A$}}
\def\bI{\mbox{\boldmath$I$}}

\def\bx{\mbox{\boldmath$x$}}

\def\bQ{\mbox{\boldmath$Q$}}

\title{\LARGE \bf
Stability Principle Underlying Passive Dynamic Walking of\\
Rimless Wheel
}
\author{Fumihiko Asano
\thanks{F. Asano is with the School of Information Science, Japan
Advanced Institute of Science and Technology, 1-1 Asahidai, Nomi,
Ishikawa 923-1292, Japan {\tt\small fasano@jaist.ac.jp}}
}

\begin{document}

\maketitle
\thispagestyle{empty}
\pagestyle{empty}
\setlength{\arraycolsep}{2pt}
\begin{abstract}
\input{abst}
\end{abstract}
\input{sec1}
\input{sec2}
\input{sec4}
\input{sec3}
\input{sec6}
\input{sec7}

\end{document}

%% file: abst.tex
Rimless wheels are known as the simplest model for passive dynamic
 walking. It is known that the passive gait generated only by gravity
 effect always becomes asymptotically stable and 1-period because a
 rimless wheel automatically achieves the two necessary conditions for
 guaranteeing the asymptotic stability; one is the constraint on impact
 posture and the other is the constraint on restored mechanical
 energy. The asymptotic stability is then easily shown by the recurrence
 formula of kinetic energy. There is room, however, for further research
 into the inherent stability principle. In this paper, we reconsider the
 stability of the stance phase based on the linearization of the
 equation of motion, and investigate the relation between the stability
 and energy conservation law. Through the mathematical analysis, we
 provide a greater understanding of the inherent stability principle.

%% file: sec1.tex
\section{INTRODUCTION}

\begin{figure}[t]
\centering
\vspace*{2mm}
\scalebox{0.9}{
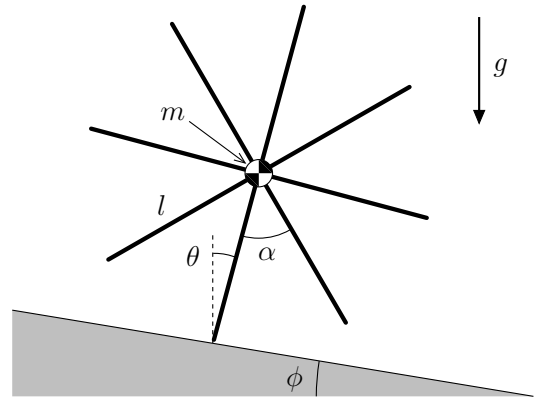
}
\caption{Passive rimless wheel model}
\label{fig2.01}
\vspace*{-3mm}
\end{figure}

A rimless spoked wheel, or simply, a rimless wheel (RW) shown in
Fig. \ref{fig2.01} has been investigated as the simplest model of
passive dynamic walking \cite{McGeer}. This is an efficient, easy,
flexible locomotion system that has both properties of wheels and legged
robots. Until now, various applications have been considered such as
active use \cite{Yan,ICRA2011}, combined motion considering the
phase difference \cite{Smith,IROS2011_1}, extension to 3D
\cite{Smith2,Narukawa}, improvement of the gait efficiency by
adding feet \cite{Jiao,Clawar2010}, and efficiency analysis of
2-period gaits \cite{AR}.

The stability inherent in a passive RW has also been studied. The
authors showed that the generated passive gait of a RW always becomes
asymptotically stable and 1-period because it automatically achieves the
two necessary conditions for asymptotic stability; one is the constraint
on impact posture and the other is the constraint on restored mechanical
energy \cite{Robotica}. The former is the condition for maintaining the
energy-loss coefficient constant; this keeps the convergence speed of
the generated gait at a constant rate. In active walkers, this can be
achieved by falling down as a 1-DOF rigid body
\cite{AR,Robotica}. The latter is not easy to be met in
underactuated limit cycle walkers with free ankle joints. Guaranteeing
the asymptotic stability rests on the problem of how to keep the
restored mechanical energy constant.

On the other hand, the simplest walking model (compass walker with
mass-less legs) also has a simplified dynamics and stability property
similar to that of RWs. If the simplest walker exhibits passive dynamic
walking while maintaining the relative hip-angle immediately prior to impact
constant of $\alpha$ [rad], the eigenvalues of the 2D Poincar\'{e}
return map becomes $\cos^2 \alpha$ and zero \cite{Wisse}. This is
because the simplest passive walker with a constant hip angle at impact
discretely behaves in the same manner as the passive RW. It is thus
considered that the eigenvalue $\cos^2 \alpha$ indicates the stability
property of a passive RW.

It is considered that the stability principle underlying passive RWs is
obvious as the author clarified using the recurrence formula of kinetic
energy \cite{Robotica}, and that there is no room for further research
into deeper understanding. In this paper, however, we synthetically
reconsider the inherent stability of a passive RW based on the above theoretical
results and show several important properties of the stance phase
undiscovered. Especially, the importance of mechanical energy and its
conservation law corresponding to the linearized dynamic equation is
suggested. Furthermore, we discuss the steady step period that is
uniquely determined in accordance with the slope and relative hip-angle.

%% file: rw.pdf_t
\begin{picture}(0,0)%
\includegraphics{rw.pdf}%
\end{picture}%
\setlength{\unitlength}{3947sp}%
\begin{picture}(3624,2756)(1189,-3373)
\put(4576,-1096){\makebox(0,0)[b]{\smash{\fontsize{12}{14.4}\usefont{T1}{ptm}{m}{n}$g$}}}
\put(3151,-3286){\makebox(0,0)[b]{\smash{\fontsize{12}{14.4}\usefont{T1}{ptm}{m}{n}$\phi$}}}
\put(2961,-2412){\makebox(0,0)[b]{\smash{\fontsize{12}{14.4}\usefont{T1}{ptm}{m}{n}$\alpha$}}}
\put(2232,-2097){\makebox(0,0)[b]{\smash{\fontsize{12}{14.4}\usefont{T1}{ptm}{m}{n}$l$}}}
\put(2450,-2451){\makebox(0,0)[b]{\smash{\fontsize{12}{14.4}\usefont{T1}{ptm}{m}{n}$\theta$}}}
\put(2310,-1435){\makebox(0,0)[b]{\smash{\fontsize{12}{14.4}\usefont{T1}{ptm}{m}{n}$m$}}}
\end{picture}%

%% file: sec2.tex
\section{INHERENT STABILITY OF RIMLESS WHEEL GAIT}

\subsection{Poincar\'{e} Return Map}

Let the subscript ``$i$'' be the number of steps. We assume that the
superscripts ``$-$'' and ``$+$'' denote immediately before and
immediately after impact. Furthermore, we assume the subscript ``eq''
and the superscript ``$*$'' express that the parameters are the steady
values.

It has already been known that the generated passive gait of a RW always
becomes $1$-period asymptotically stable because it always falls down as
a 1-DOF rigid body and the energy-loss coefficient and restored
mechanical energy are automatically kept constant. In the following, we
outline the stability principle.

Let $K_i^-$ [J] be the kinetic energy immediately before the $i$-th
impact. Then the following recurrence formula holds.
\begin{equation}
K_{i+1}^- = \varepsilon K_i^- + \Delta E
\label{eq2.01}
\end{equation}
Where $\varepsilon$ [-] is the energy-loss coefficient and $\Delta E$
[J] is the restored mechanical energy by gravity. They are respectively
given by
\begin{eqnarray}
\varepsilon &=& \cos^2 \alpha, \\
\Delta E &=& 2 m g l \sin \frac{\alpha}{2} \sin \phi,
\label{eq2.02}
\end{eqnarray}
and are positive constants. Following Eq. (\ref{eq2.01}), $K_i^-$
converges to
\begin{equation}
K_{\rm eq}^- := \lim_{i \rightarrow \infty} K_i^-
= \frac{\Delta E}{1 - \varepsilon}.
\label{eq2.03}
\end{equation}
This gives the proof of the asymptotic stability. We can rearrange
Eq. (\ref{eq2.01}) considering the error term as
\begin{equation}
K_{\rm eq}^- + \Delta K_{i+1}^- = \varepsilon 
\left(
K_{\rm eq}^- + \Delta K_i^-
\right)
+ \Delta E,
\label{eq2.04}
\end{equation}
and subtracting the following steady equation
\begin{equation}
K_{\rm eq}^- = \varepsilon K_{\rm eq}^- + \Delta E
\label{eq2.05}
\end{equation}
from Eq. (\ref{eq2.04}), we get
\begin{equation}
\Delta K_{i+1}^- = \varepsilon 
\Delta K_i^-.
\label{eq2.06}
\end{equation}
Therefore, the convergence of kinetic energy is determined in accordance
with the energy-loss coefficient. Here, note that the following relation holds.
\begin{eqnarray}
K_i^- &=& \frac{1}{2} m l^2 
\left(
\dth_i^- 
\right)^2
= 
\frac{1}{2} m l^2 \left(
\dth_{\rm eq}^- + \Delta \dth_i^-
\right)^2 \nonumber \\
&\approx&
\frac{1}{2} m l^2 \left(
\dth_{\rm eq}^-
\right)^2 + m l^2 \dth_{\rm eq}^- \Delta \dth_i^- \nonumber \\
&=& K_{\rm eq}^- + m l^2 \dth_{\rm eq}^- \Delta \dth_i^-
\label{eq2.07}
\end{eqnarray}
This leads to
\begin{equation}
\Delta K_i^- = K_i^- - K_{\rm eq}^- = 
m l^2 \dth_{\rm eq}^- \Delta \dth_i^-,
\label{eq2.08}
\end{equation}
and by substituting this into Eq. (\ref{eq2.06}) we get
\begin{equation}
\Delta \dth_{i+1}^- = \varepsilon \Delta \dth_{i}^-.
\label{eq2.09}
\end{equation}
The angular velocity also converges to the steady value in the same
manner as kinetic energy.

\subsection{Stability of Collision Phase}

The relation of angular velocity immediately before impact and that
immediately after impact is given by
\begin{equation}
\dth_i^+ = \cos \alpha \cdot \dth_i^-.
\label{eq2.10}
\end{equation}
We can rearrange this considering the error term as
\begin{equation}
\dth_{\rm eq}^+ + \Delta \dth_{i}^+ = 
\cos \alpha 
\left(
\dth_{\rm eq}^- + \Delta 
\dth_i^-
\right).
\label{eq2.11}
\end{equation}
By subtracting the following steady equation
\begin{equation}
\dth_{\rm eq}^+ = 
\cos \alpha \cdot
\dth_{\rm eq}^- 
\label{eq2.12}
\end{equation}
from Eq. (\ref{eq2.11}), we get
\begin{equation}
\Delta \dth_{i}^+ = 
\cos \alpha \cdot
\Delta \dth_i^-.
\label{eq2.13}
\end{equation}
In the following, we denote $\bar{R} = \cos \alpha$. In addition,
following Eq. (\ref{eq2.12}), this is also specified as the ratio of the
steady angular velocities:
\begin{equation}
\bar{R} = \frac{
\dth_{\rm eq}^+
}{
\dth_{\rm eq}^- 
}.
\label{eq2.14}
\end{equation}

\subsection{Stability of Stance Phase}

Let us assume that the transition function of the state error during the
stance phase is specified as
\begin{equation}
\Delta \dth_{i+1}^- = 
\bar{Q}
\Delta \dth_i^+.
\label{eq2.145}
\end{equation}
Following Eqs. (\ref{eq2.09}) and (\ref{eq2.13}), the transition function, $\bar{Q}$, is then solved as
\begin{equation}
\bar{Q} = \frac{\varepsilon}{\bar{R}} = \cos \alpha.
\label{eq2.15}
\end{equation}
Therefore, we can find that the transition of the state error during the
stance and collision phases are identical and are $\cos \alpha$. In the
subsequent sections, we will investigate this result in more detail from
the mechanical energy point of view.

%% file: sec4.tex
\section{LINEARIZATION OF MOTION AND MECHANICAL ENERGY}

\subsection{Linearization of Dynamic Equation and State Space Realization}

The dynamic equation of the RW shown in Fig. \ref{fig2.01} is given by
\begin{equation}
m l^2 \ddot{\theta} - m g l \sin \theta = 0,
\label{eq4.01}
\end{equation}
and its linearization around $\theta = \dth = 0$ becomes
\begin{equation}
m l^2 \ddot{\theta} - m g l \theta = 0.
\label{eq4.015}
\end{equation}
This can be arranged as
\begin{eqnarray}
\ddot{\theta} &=& \omega^2 \theta, \\
\label{eq4.02}
\omega &:=& \sqrt{\frac{g}{l}}.
\end{eqnarray}
The state space representation then becomes
\begin{equation}
\frac{\rm d}{{\rm d}t}
\left[
\begin{array}{c}
\theta \\
\dth
\end{array}
\right] = 
\left[
\begin{array}{cc}
0 & 1 \\
\omega^2 & 0
\end{array}
\right]
\left[
\begin{array}{c}
\theta \\
\dth
\end{array}
\right].
\label{eq4.03}
\end{equation}
In the following, we denote this as $\dot{\bx} = \bA \bx$.

\subsection{Linearized Mechanical Energy}

The author investigated the inherent self-stabilization principle of a
passive compass-gait by introducing the linearized mechanical energy and
numerically showed the potentiality that the passive compass gait is
stabilized not by means of the state variables but by means of the
mechanical energy \cite{IROS2011_2}. In this paper, we also use it for
analysis of a passive RW.

The kinetic energy of the original (nonlinear) RW is determined as
\begin{equation}
K (\dth) = \frac{1}{2} m l^2 \dth^2,
\label{eq3.005}
\end{equation}
and the corresponding one to the linearized model is identical to
this. Whereas the potential energy for the original RW is determined as
$P (\theta) = m g l \cos \theta$ [J], and we consider its second-order
approximation
\begin{equation}
P(\theta) = m g l \left(
1 - \frac{\theta^2}{2}
\right),
\label{eq3.01}
\end{equation}
Let us define Eq. (\ref{eq3.01}) as the potential energy corresponding
to the linearized system in the sense that this leads to the dynamic
equation of Eq. (\ref{eq4.015}) together with $K (\dth)$. In addition,
let us define the maximum potential energy, $P_{\rm max} := m g l$ [J];
this is the maximum potential energy the robot can reach during the
stance phases. Eq. (\ref{eq3.01}) is then rewritten as follows.
\begin{equation}
P(\theta) = P_{\rm max} - \frac{1}{2} m g l \theta^2
\label{eq3.02}
\end{equation}
The total mechanical energy is then determined as
\begin{equation}
E (\bx) = P_{\rm max} + \frac{1}{2} \bx^{\rm T} \bW_0 \bx,
\label{eq3.03}
\end{equation}
where
\begin{equation}
\bW_0 := \left[
\begin{array}{cc}
- m g l & 0 \\
0 & m l^2
\end{array}
\right]
\label{eq3.04}
\end{equation}
is a constant matrix including the inertia and gravity. 

The following equation derived by using Eqs. (\ref{eq3.005}) and
(\ref{eq3.01}) becomes identical to the linearized dynamic equation
(\ref{eq4.015}).
\begin{equation}
\frac{\rm d}{{\rm d}t} \frac{\partial K (\dth)}{\partial \dth} - 
\frac{\partial K (\dth)}{\partial \theta} + \frac{\partial P
(\theta)}{\partial \theta} = 0
\end{equation}
Where $
\partial K (\dth)/\partial \theta = 0
$.
This shows that Eq. (\ref{eq3.03}) succeeds the
conservation property of mechanical energy from nonlinear to linearized
system.

The time-derivative of $E (\bx)$ becomes
\begin{equation}
\frac{{\rm d}E (\bx)}{{\rm d}t} = \bx^{\rm T} \bW_0 \dot{\bx} = \bx^{\rm
 T} \bW_0 \bA \bx.
\label{eq3.05}
\end{equation}
Here, the product $\bW_0 \bA$ has the form
\begin{equation}
\bW_0 \bA = m l^2 \left[
\begin{array}{cc}
0 & - \omega^2 \\
\omega^2 & 0
\end{array}
\right]
\label{eq3.06}
\end{equation}
and is skew-symmetric. Therefore, the value of Eq. (\ref{eq3.05}) always
becomes zero and this shows that the total mechanical energy is kept
constant also in the linearized system.

\subsection{Linearized Motion and Its Effective Range}

Let the contact point with the ground be the origin, the CoM position of
the nonlinear model, $(x_c,z_c)$, then becomes $(l \sin \theta, l \cos
\theta)$ and it moves along a circular orbit. In the following, we
discuss how the CoM position of the linearized model,
$(\bar{x}_c,\bar{z}_c)$, moves.

Since the potential energy of the linearized model is given by
Eq. (\ref{eq3.01}), the $Z$-position of it, $\bar{z}_c$, can be
determined as
\[
 \bar{z}_c = l \left(
1 - \frac{\theta^2}{2}
\right).
\]
Its time derivative becomes
\[
 \dot{\bar{z}}_c = - l \theta \dth.
\]
The kinetic energy is also given by
\[
 K = \frac{1}{2} m \left(
\dot{\bar{x}}^2_c + \dot{\bar{z}}^2_c
\right) = \frac{1}{2} m \left(
\dot{\bar{x}}^2_c + l^2 \theta^2 \dth^2
\right),
\]
and this must be equal to Eq. (\ref{eq3.005}). The time derivative of
the $X$-position of the CoM, $\bar{x}_c$, can be therefore determined as
\[
\dot{\bar{x}}_c = l \dth \sqrt{1 - \theta^2},
\]
and $\bar{x}_c$ is also able to be obtained by integrating this with
respect to time as follows.
\[
 \bar{x}_c = \int_{0}^{\theta} 
 l \sqrt{1 - x^2} \ {\rm d}x = 
\frac{1}{2} l \left(
\theta \sqrt{1 - \theta^2} + \sin^{-1} \theta
\right)
\]
The range of $\theta$ is, however, $\left|
\theta
\right| \leq 1$. The maximum of $\left|
\theta
\right|$ is $\theta_{\rm eq}^-$, and we therefore define the effective
range of linearization as
\begin{equation}
\max \left|
\theta
\right| = 
\theta_{\rm eq}^- = \frac{\alpha}{2} + \phi \leq 1.
\label{eq3.05}
\end{equation}

\begin{figure}[!b]
\centering
\vspace*{-2mm}
\includegraphics[width=1.0\linewidth]{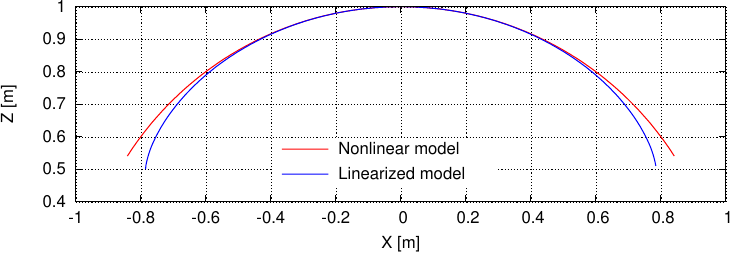}\\
\vspace*{-2mm}
\caption{Trajectories of center of mass positions of nonlinear and linearized models}
\label{fig3.01}
\end{figure}

Fig. \ref{fig3.01} plots $(x_c,z_c)$ and $(\bar{x}_c,\bar{z}_c)$ for
comparison. The CoM of the linearized model moves along a convex curve
that is near-circular orbit but the curvature is not constant. We can
understand from the above that the linearized equation of motion
(\ref{eq4.015}) is a feasible revolving motion that satisfies the law of
conservation of mechanical energy.

\subsection{Numerical Example}

\begin{figure}[!t]
\vspace*{2mm}
\centering
\includegraphics[width=1.0\linewidth]{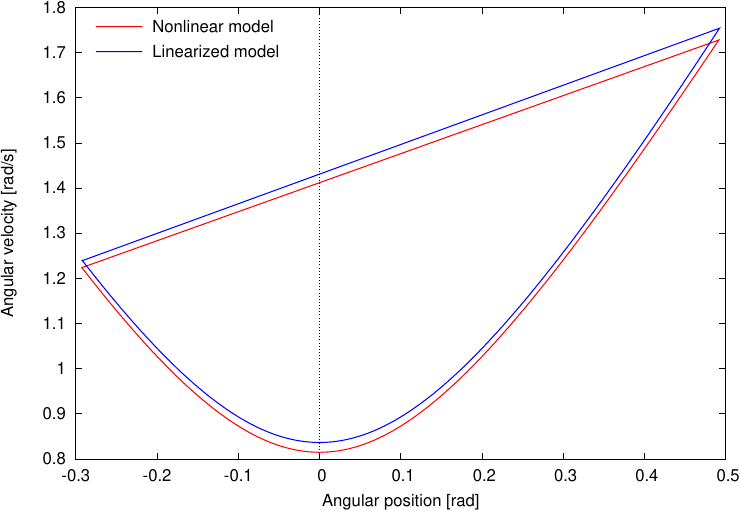}\\
\vspace*{-2mm}
\caption{Trajectories of steady gaits of nonlinear and linearized models
 in phase space}
\label{fig3.02}
\bigskip
\centering
\includegraphics[width=1.0\linewidth]{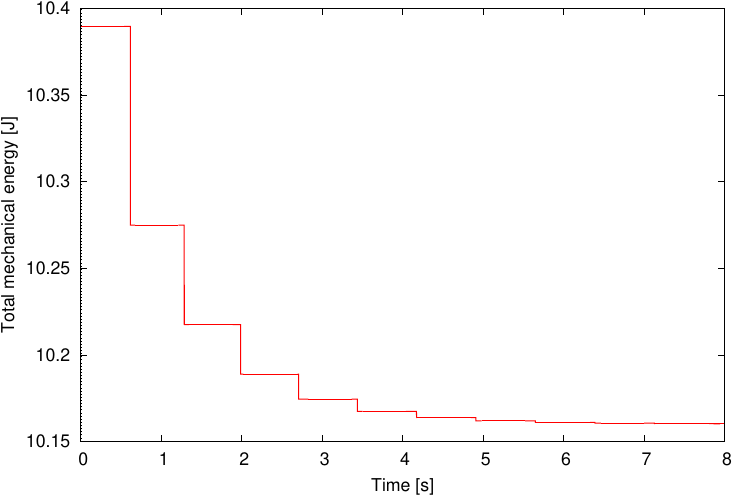}\\
\vspace*{-2mm}
\caption{Time evolution of total mechanical energy in linearized passive-dynamic gait}
\label{fig3.03}
\bigskip
\centering
\includegraphics[width=1.0\linewidth]{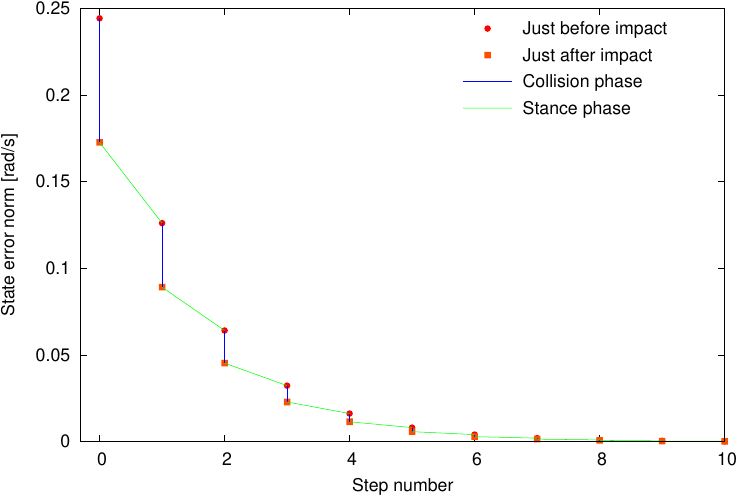}\\
\vspace*{-2mm}
\caption{Evolution of state error with respect to step number in passive-dynamic gait of Fig. \ref{fig3.03}}
\label{fig3.04}
\end{figure}

Fig. \ref{fig3.02} shows the trajectories of the generated steady
passive-dynamic gaits of the nonlinear and linearized models where $\phi
= 0.10$ [rad]. We can confirm that both trajectories are similar and the
linear approximation is thus valid. Fig. \ref{fig3.03} shows the time
evolution of the linearized mechanical energy, $E(\bx)$. The RW started
passive dynamic walking from the impact posture with a high angular
velocity. We can see that the value is kept constant during the stance
phases and discontinuously changes at the collision phases. The
conservation property is represented.

Fig. \ref{fig3.04} plots the evolution of the state errors, $ \left|
\Delta \dth_i^- \right|$ and $\left| \Delta \dth_i^+ \right|
$, with respect to the step number corresponding to
Fig. \ref{fig3.03}. We can see that, as explained in Section II, the
error norm monotonically decreases during both phases at a constant rate
of $\cos \alpha$. It finally converges to zero with the convergence of
the linearized mechanical energy in Fig. \ref{fig3.03}.

%% file: sec3.tex
\section{STABILITY ANALYSIS OF STANCE PHASE}

\subsection{Transition Function for State Error of Stance Phase}

The relation between the state vector immediately before the $(i+1)$-th
impact and that immediately after the $i$-th impact is given by
\begin{equation}
\bx_{i+1}^- = \rme^{\bAs T_i} \bx_i^+.
\label{eq4.04}
\end{equation}
In a steady gait, the following equation holds.
\begin{equation}
\bx_{\rm eq}^- = \rme^{\bAs T^*} \bx_{\rm eq}^+
\label{eq4.05}
\end{equation}
This equation is detailed as
\begin{equation}
\left[
\begin{array}{c}
\theta_{\rm eq}^- \\
\dth_{\rm eq}^-
\end{array}
\right] = 
\left[
\begin{array}{cc}
\cosh (\omega T^*) & \omega^{-1} \sinh (\omega T^*) \\
\omega \sinh (\omega T^*) & \cosh (\omega T^*)
\end{array}
\right]
\left[
\begin{array}{c}
\theta_{\rm eq}^+ \\
\dth_{\rm eq}^+
\end{array}
\right],
\label{eq4.055}
\end{equation}
where $T^*$ [s] is the steady step period.

By expanding Eq. (\ref{eq4.04}) using the approximation of 
$\rme^{\bAs \Delta T_i} \approx \bI_2 + \bA \Delta T_i$ and eliminating
the terms higher than second order, we get
\begin{equation}
\bx_{i+1}^- = \bx_{\rm eq}^- + \rme^{\bAs T^*} \Delta \bx_i^+ + \bA
 \bx_{\rm eq}^- \Delta T_i.
\label{eq4.06}
\end{equation}
Here, let us define a projection vector $\bp := \left[
\begin{array}{cc}
1 & 0
\end{array}
\right]$. By premultiplying the state vector by $\bp$, we get
\begin{equation}
\bp \bx_i^- = \bp \bx_{\rm eq}^- = \theta_{\rm eq}^- = \frac{\alpha}{2} + \phi
\end{equation}
Considering this and multiplying both sides of Eq. (\ref{eq4.06}) by
$\bp$, we get
\[
 0 = \bp \rme^{\bAs T^*} \Delta \bx_i^+ + \bp \bA \bx_{\rm eq}^- \Delta T_i.
\]
Then we can solve this for the error of the steady step period, $\Delta
T_i$, as follows.
\begin{equation}
\Delta T_i = - \frac{
\bp 
\rme^{\bAs T^*} \Delta \bx_i^+
}{
\bp \bA \bx_{\rm eq}^-
}
\end{equation}
By substituting this into Eq. (\ref{eq4.06}), we can get the transition
function for the state error of the stance phase, $\bQ \in \mR^{2 \times
2}$, as follows.
\begin{eqnarray}
\Delta \bx_{i+1}^- &=& \bQ \Delta \bx_{i}^+, 
\\ 
\bQ &:=& \left(
\bI_2 - \frac{\bA \bx_{\rm eq}^- \bp}{\bp \bA \bx_{\rm eq}^-}
\right) \rme^{\bAs T^*}
\end{eqnarray}
Unlike the passive compass gait without constraint on impact posture
\cite{ICRA2011_2}, in limit cycle walking with the constraint, the
errors of the angular positions are always zero and we can reduce the
redundancy of $\bQ$ \cite{Hu2011}. In passive dynamic walking of a RW,
the following relation holds.
\begin{equation}
\Delta \bx_{i}^\pm = 
\left[
\begin{array}{c}
0 \\
1
\end{array}
\right]
\Delta \dth_{i}^\pm
\end{equation}
$\bQ$ can be reduced to
\begin{equation}
\Delta \dth_{i+1}^- = \bar{Q} \Delta \dth_{i}^+ ,
\end{equation}
where
\begin{eqnarray}
\bar{Q} &:=& 
\left[
\begin{array}{c}
0 \\
1
\end{array}
\right]^{\rm T} \bQ
\left[
\begin{array}{c}
0 \\
1
\end{array}
\right] \nonumber \\
&=& 
\cosh (\omega T^*) - \frac{\theta_{\rm eq}^- \omega}{\dth_{\rm eq}^-}
\sinh (\omega T^*).
\label{eq4.07}
\end{eqnarray}
Note that $\dth_{\rm eq}^- \neq 0$ must hold in this equation. It is
obvious, however, that $\dth_{\rm eq}^- > 0$ is always achieved in a
stable walking. We therefore assume that this condition always holds in
the following.

\subsection{Simplifications of $\bar{Q}$ and Poincar\'{e} Return Map}

In the following, we reconsider the physical meaning of $\bar{Q}$ and
derive another formulation of it without including $T^*$.

Eq. (\ref{eq4.055}) can be rewritten as
\begin{equation}
\left[
\begin{array}{c}
\theta_{\rm eq}^- \\
\dth_{\rm eq}^-
\end{array}
\right] = 
\left[
\begin{array}{cc}
\dth_{\rm eq}^+ \omega^{-1} & \theta_{\rm eq}^+ \\
\theta_{\rm eq}^+ \omega & \dth_{\rm eq}^+
\end{array}
\right]
\left[
\begin{array}{c}
\sinh (\omega T^*) \\
\cosh (\omega T^*)
\end{array}
\right].
\label{eq4.081}
\end{equation}
$\sinh (\omega T^*)$ and $\cosh (\omega T^*)$ are then solved as
\begin{eqnarray}
\hspace*{-10mm}
 & & \left[
\begin{array}{c}
\sinh (\omega T^*) \\
\cosh (\omega T^*)
\end{array}
\right] = 
\left[
\begin{array}{cc}
\dth_{\rm eq}^+ \omega^{-1} & \theta_{\rm eq}^+ \\
\theta_{\rm eq}^+ \omega & \dth_{\rm eq}^+
\end{array}
\right]^{-1}
\left[
\begin{array}{c}
\theta_{\rm eq}^- \\
\dth_{\rm eq}^-
\end{array}
\right] \nonumber \\
\hspace*{-10mm}
& & = 
\frac{
\omega
}{
\left(
\dth_{\rm eq}^+
\right)^2 -
\left(
\theta_{\rm eq}^+
\right)^2 \omega^2
} \left[
\begin{array}{c}
\dth_{\rm eq}^+ \theta_{\rm eq}^- - \theta_{\rm eq}^+ 
\dth_{\rm eq}^-
\\
\dth_{\rm eq}^+ \dth_{\rm eq}^- \omega^{-1}
- \theta_{\rm eq}^+ \theta_{\rm eq}^- \omega
\end{array}
\right]
\label{eq4.082}
\end{eqnarray}
By substituting this into Eq. (\ref{eq4.07}), we get
\begin{equation}
\bar{Q} = \frac{\dth_{\rm eq}^+}{\dth_{\rm eq}^-} \cdot
\frac{
\left(
\dth_{\rm eq}^-
\right)^2 - 
\left(
\theta_{\rm eq}^-
\right)^2 \omega^2
}{
\left(
\dth_{\rm eq}^+
\right)^2 - 
\left(
\theta_{\rm eq}^+
\right)^2 \omega^2
}.
\label{eq4.08}
\end{equation}
Here, considering the relation
\begin{eqnarray}
\dth^2 - 
\theta^2 \omega^2 &=& 
\frac{2}{ml^2}
\left(
\frac{1}{2} m l^2 \dth^2 - 
\frac{1}{2} m g l \theta^2
\right) \nonumber \\
&=& 
\frac{2}{ml^2}
\left(
E - P_{\rm max}
\right),
\label{eq4.09}
\end{eqnarray}
Eq. (\ref{eq4.08}) can be rewritten as
\begin{equation}
\bar{Q} = \frac{\dth_{\rm eq}^+}{\dth_{\rm eq}^-} \cdot
\frac{E_{\rm eq}^- - P_{\rm max}}{E_{\rm eq}^+ - P_{\rm max}}.
\label{eq4.10}
\end{equation}
In a passive-dynamic gait, the law of conservation of mechanical energy,
$E_{\rm eq}^+ = E_{\rm eq}^-$, holds. We then get the following
equation.
\begin{equation}
\frac{E_{\rm eq}^- - P_{\rm max}}{E_{\rm eq}^+ - P_{\rm max}} = 1
\label{eq4.11}
\end{equation}
Therefore, $\bar{Q}$ finally becomes
\begin{equation}
\bar{Q} = \frac{\dth_{\rm eq}^+}{\dth_{\rm eq}^-}.
\label{eq4.12}
\end{equation}
This does not depend on the steady step period. In addition, this is
equal to $\cos \alpha$; the stability property is determined only by the
configuration. Note also that both the numerator and denominator in
Eq. (\ref{eq4.11}) are positive because they are energy margins
necessary for overcoming the potential barrier at mid-stance. They must
be positive in stable passive-dynamic gaits.

Following Eqs. (\ref{eq2.14}) and (\ref{eq4.12}), the Poincar\'{e}
return map can be formulated as
\begin{equation}
\bar{Q} \bar{R} = 
\bar{R} \bar{Q} = 
\frac{
\left(
\dth_{\rm eq}^+
\right)^2
}{
\left(
\dth_{\rm eq}^-
\right)^2
} = \cos^2 \alpha.
\label{eq4.13}
\end{equation}
It is also obvious that this is equal to $\cos^2 \alpha$. This is
identical to the eigenvalue of the simplest walking model that falls
down with a constant hip angle \cite{Wisse}.

%% file: sec6.tex
\section{STEADY STEP PERIOD}

Following Eq. (\ref{eq4.082}), we get 
\begin{equation}
\tanh \left(
\omega T^*
\right) = F (\alpha, \phi),
\label{eq6.01}
\end{equation}
where
\begin{eqnarray}
F (\alpha, \phi) &=& \frac{
\dth_{\rm eq}^+ \theta_{\rm eq}^- - \theta_{\rm eq}^+ 
\dth_{\rm eq}^-
}{
\dth_{\rm eq}^+ \dth_{\rm eq}^- \omega^{-1}
- \theta_{\rm eq}^+ \theta_{\rm eq}^- \omega
} \nonumber \\
&=& \frac{2 \sin \alpha \sqrt{2 \alpha \phi}}{
\alpha \left(
1 - \cos \alpha
\right) + 2 \phi \left(
1 + \cos \alpha
\right)
}.
\label{eq6.02}
\end{eqnarray}
Since both the numerator and denominator of Eq. (\ref{eq6.02}) are
positive, $F (\alpha, \phi) > 0$ holds. In addition, following the
relation between arithmetic and geometric means, the maximum value
becomes
\[
 F (\alpha, \phi) \leq 
\frac{2 \sin \alpha \sqrt{2 \alpha \phi}}{
2\sqrt{
2 \alpha \phi \left(
1 - \cos \alpha
\right)
\left(
1 + \cos \alpha
\right)
}
} = 1,
\]
where the equality holds in the following case.
\begin{eqnarray*}
 & & \alpha \left(
1 - \cos \alpha
\right) = 2 \phi \left(
1 + \cos \alpha
\right) \nonumber \\
 & & \Longleftrightarrow \ \phi = \frac{\alpha}{2} \tan^2
\frac{\alpha}{2} =: \phi_{\min} 
\end{eqnarray*}
Furthermore, the derivative of $F (\alpha, \phi)$ with respect to $\phi$
becomes
\begin{eqnarray*}
 \frac{\partial F (\alpha, \phi)}{\partial \phi} &=& 
\sqrt{
\frac{2 \alpha}{\phi}
} \sin \alpha \nonumber \\
 & & \times \frac{
\alpha \left(
1 - \cos \alpha
\right) - 2 \phi \left(
1 + \cos \alpha
\right)
}{
\left(
\alpha
\left(
1 - \cos \alpha
\right) + 
2 \phi
\left(
1 - \cos \alpha
\right)
\right)^2
}.
\end{eqnarray*}
The condition that this becomes zero is identical to the condition that
the numerator becomes zero, and this is solved as follows.
\[
 \frac{\partial F (\alpha, \phi)}{\partial \phi} = 0 \ 
\Longleftrightarrow \ 
\phi = \frac{\alpha}{2} \tan^2 \frac{\alpha}{2} = \phi_{\min}
\]
In addition, $F (\alpha, 0) = 0$ holds for all $\alpha$. We can conclude
from the above that $F (\alpha, \phi)$ monotonically increases from $0$
to $1$ in the range of $0 \leq \phi \leq \phi_{\min}$, and that it
monotonically decreases in $\phi > \phi_{\min}$. Eq. (\ref{eq6.01})
therefore has a unique solution
\begin{equation}
T^* = \frac{1}{\omega} \tanh^{-1} F (\alpha, \phi)
\end{equation}
in $\phi \geq \phi_{\min}$.

\begin{figure}[t]
\vspace*{3mm}
\includegraphics[width=1.0\linewidth]{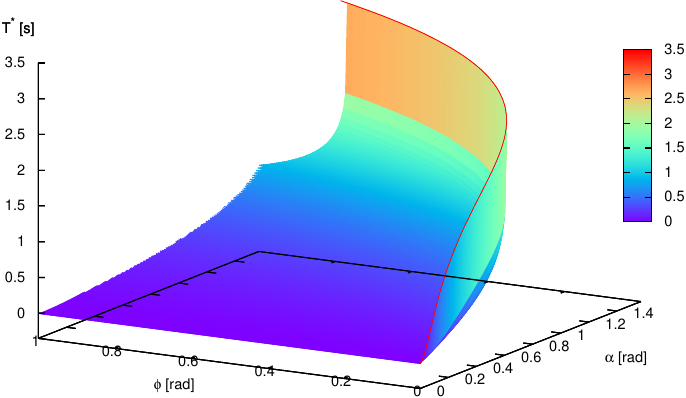}
\caption{3D plot of $T^* (\alpha, \phi)$}
\label{fig5.01}
\end{figure}

Fig. \ref{fig5.01} plots $T^*$ as a function of $\alpha$ and $\phi$ in
the domain of $\phi_{\min} < \phi < 1.0$ and $0 < \alpha < \pi/2$. Since
$T^* (\alpha, \phi)$ diverges as $\phi \rightarrow \phi_{\min}$, we
chose the minimum value of $\phi$ as $\phi_{\min} + \epsilon$ where
$\epsilon = 0.0001$. We also plot the curve of $T^* (\alpha, \phi_{\min}
+ \epsilon)$ as a red line in the figure for viewability. $T^*$
drastically increases in the neighborhood of $\phi_{\min}$. This is
because the RW marginally overcomes the potential barrier at mid-stance
in this case.

\begin{figure}[t]
\includegraphics[width=1.0\linewidth]{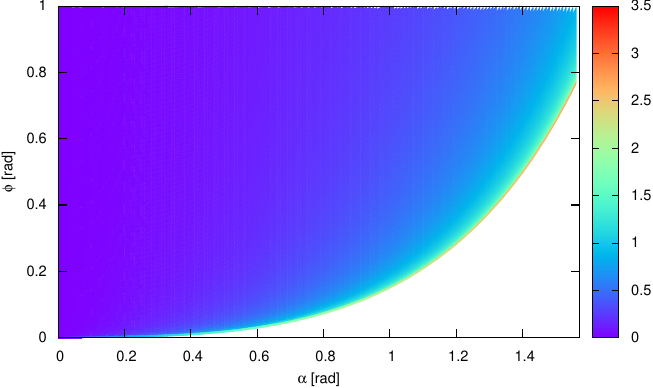}\\
\vspace*{-6mm}
\caption{2D plot of $T^* (\alpha, \phi)$}
\label{fig5.02}
\vspace*{-2mm}
\end{figure}

Fig. \ref{fig5.02} shows the 2D plot of Fig. \ref{fig5.01} in
$\alpha$-$\phi$ plane. From the figure, we can understand that the
steady step period is less than $1.5$ [s] in almost every case of
$(\alpha,\phi)$. It is obvious that the steady step period monotonically
decreases as the slope increases because the walking speed increases. We
must consider, however, that the condition of unilateral constraint
should be guaranteed during the stance phases. RWs begin to jump as the
slope becomes steep, that is, the generated motion changes from walking
to running or skipping. The domain of Figs. \ref{fig5.01} and
\ref{fig5.02} is therefore redundant. It is true that RWs must have
sufficient kinetic energy just after impact for overcoming the potential
barrier, and $\phi_{\min}$ indicates the lower bound of $\phi$. We
should identify the upper bound of $\phi$ that indicates the condition
of unilateral constraint in the future.

%% file: sec7.tex
\section{CONCLUSION AND FUTURE WORK}

In this paper, we investigated the stability principle inherent in
passive dynamic walking of a RW. We introduced mechanical energy
corresponding to the linearized model and mathematically clarified the
relation between the stability of stance phase and mechanical energy. We
also analytically derived the steady step period as a function of
$\alpha$ and $\phi$, and identified the lower bound of $\phi$ which is
identical to the condition that the RW can overcome the potential
barrier at mid-stance.

As mentioned in the previous section, we should identify the upper bound
of $\phi$ and specify the forming condition of passive dynamic walking
of a RW in the future. Comparison of the results with those of the
nonlinear model is also necessary.

The greatest contribution of this paper was to develop the fundamental
method for understanding the stability principle underlying limit cycle
walking with constraint on impact posture. The stability of active RWs
and bipedal walkers that fall down as a 1-DOF rigid body in the same
posture is being explained in the same manner as the passive RW; their
transition functions for the stance phase can be written in the same
form of Eq. (\ref{eq4.12}). The results will be reported in a future
paper.
